%% file: main.tex
\documentclass{article}
\usepackage[preprint]{neurips_2026}
\usepackage[utf8]{inputenc}
\usepackage[T1]{fontenc}
\usepackage{courier} 
\usepackage{booktabs,graphicx,amsmath,amssymb,microtype,xcolor}
\usepackage[hidelinks]{hyperref}

\title{Geometry Conditioning in an Embodied SLM:\\
Training Controls and Robustness Diagnostics in a 0.8B Hybrid Model}

\author{%
  Hao Li\thanks{Equal contribution.}\\
  \texttt{thomasli114514@gmail.com}
  \And
  Haofei Sun\footnotemark[1]\\
  \texttt{humphreysun98@gmail.com}
  \AND
  Lin He\thanks{Corresponding author.}\\
  University of Tennessee, Knoxville\\
  \texttt{lynnhe@utk.edu}
}

\begin{document}
\maketitle

\begin{abstract}
We study how physical-state inputs affect a 0.8B hybrid language model
adapted for manipulation with 6.2M trainable parameters. Six conditions are
trained on three LIBERO-Spatial tasks and evaluated over three seeds and
540 held-out rollouts. Conditioning recurrent decay gates on geometric
increments yields 28.9\% success, compared with 36.7\% when those increments
are shuffled during training and 24.4\% without explicit object/goal geometry.
Both geometry policies receive correct inputs at evaluation. A token adapter
using the same increments scores 27.8\%; differences vary across seeds and
remain inconclusive. Token-clock conditioning scores 11.1\%, including one
seed that fails to converge. In separate robustness tests, a state-only
relative-coordinate policy retains 7/10 success under frame relabeling,
whereas all four tested visual policies fall to at most 3/20 after a 5\,cm
object displacement. These results show no reliable advantage from
training-time geometric alignment under this recipe and illustrate the gap
between coordinate invariance and physical-layout generalization. Episode
records, seed-level analyses, and figure-generation code accompany the paper.
\end{abstract}

\section{Introduction}
\input{sections/intro}

\section{Setup: one frozen SLM, two pathways}
\input{sections/setup}

\section{Findings}
\input{sections/findings}

\section{Implications and limitations}
\input{sections/discussion}

\bibliographystyle{plainnat}
\bibliography{refs}

\appendix
\input{sections/appendix}

\end{document}

%% file: sections/intro.tex
Small language models can be adapted into robot policies with only a few
million trainable parameters. For hybrid backbones that combine attention
with gated recurrence, geometry can enter through the input tokens or the
recurrent decay gates. It is unclear whether either route helps a visual
policy learn to manipulate objects, or whether apparent gains depend on the
geometric signal being correctly aligned with the demonstrations.

We compare these routes in Qwen3.5-0.8B on
LIBERO-Spatial~\citep{liu2023libero}. The base weights are frozen; LoRA,
an input projection, and an action head are trained. A token-MLP adapter
(c1) and a decay-gate adapter (f0) receive the same six-dimensional geometric
increments at nearly equal parameter budgets. A second token adapter (c2)
receives richer relative-state features. Within the gate route, we compare
correct geometry, geometry shuffled during training, and token clocks
(Fig.~\ref{fig:architecture}).

Correct geometry does not outperform shuffled training in the recorded
rollouts, and the c1--f0 comparison is unresolved. The clock condition has
lower success and one failed training seed. We examine these differences
per seed and task, then test two kinds of robustness: relabeling coordinates
without moving the scene, and physically displacing an object. Relative
coordinates handle the first test, but the tested visual policies fail the
second. The study identifies limits of geometry conditioning under a small
adaptation budget and provides controls for interpreting its apparent benefits.

\paragraph{Related work.}
Visual manipulation policies such as OpenVLA~\citep{kim2024openvla},
OpenVLA-OFT~\citep{kim2025oft}, and ACT~\citep{zhao2023act} learn action
predictions from demonstrations. MiniVLA~\citep{minivla2024} explores smaller
language backbones, and BAKU~\citep{haldar2024baku} studies efficient multi-task
policies. RoboMamba~\citep{liu2024robomamba} applies state-space models to
robotic reasoning and manipulation; here we study a conditioning interface
in gated delta networks~\citep{yang2024gdn}.
Related applications in built environments include architectural feature
recognition~\citep{zou2023architecture} and language-guided robotic air-quality
monitoring~\citep{he2025iaq}. Our experiments concern the lower-level problem
of converting physical observations into manipulation actions.
Noise injection in imitation learning~\citep{laskey2017dart} suggests a possible
regularization role for shuffled geometry. A separate line of work on
personalized agents finds that updated stored state need not change generated
behavior~\citep{sun2026stale}; in our setting, the corresponding empirical
question is whether adding physical state changes closed-loop performance.

%% file: sections/setup.tex
\begin{figure}[t]
\centering
\includegraphics[width=\linewidth]{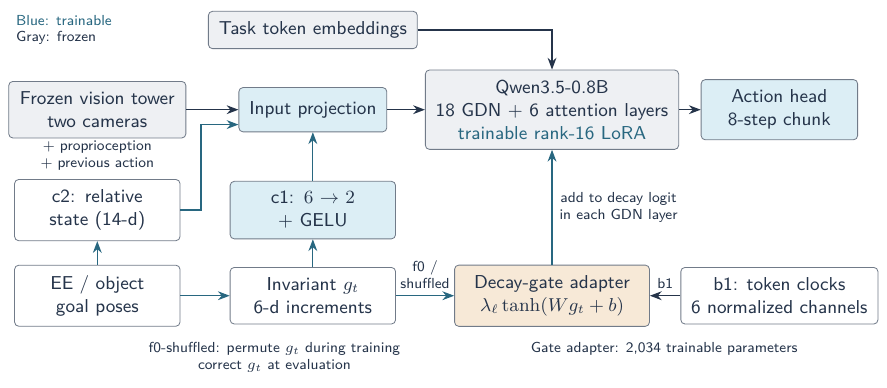}
\caption{Geometry-conditioning routes. Conditions are trained separately:
c0 omits the geometry branches; c1 and c2 add token features; f0 and
f0-shuffled condition the GDN gates; b1 supplies clock channels to the gate
adapter instead of geometry. Base weights are frozen. Blue modules, LoRA,
and the orange gate adapter are trainable.}
\label{fig:architecture}
\end{figure}

\paragraph{Host and adaptation.} We use the 0.8B hybrid LM (Qwen3.5-0.8B) with a
3:1 ratio of gated-deltanet (GDN)~\citep{yang2024gdn} layers to full-attention layers. Policies
are trained by behavior cloning on LIBERO-Spatial with action chunking~\citep{zhao2023act}
($K{=}8$), an L1 objective~\citep{kim2025oft} with gripper-event proximity weighting, previous-action noise, frozen dual-camera vision features, and language conditioning.
Trainable parameters are matched to within 0.25\% across conditions (6.215M--6.230M; Table~\ref{tab:cond}): LoRA~\citep{hu2022lora}
rank-16 on attention/QKV projections --- GDN decay projections are
deliberately excluded so that decay-gate conditioning remains exclusive to
the conditions under study --- plus a linear input projection and a 512-unit action head. Vision features are the frozen vision tower of the same checkpoint, mean-pooled to 768-d per camera (agent-view and wrist), concatenated and scaled by 0.22 into the input projection.

\paragraph{Gate conditioning.} Each of the 18 GDN layers computes a per-head
decay logit $a_{t,h}$ from the hidden state. We add a scaled, tanh-squashed bias driven by a
6-d physical feature vector $g_t$ (object and goal translation-increment norms divided by 0.5\,m,
rotation-increment angles divided by $\pi$, $\log(1{+}\Delta t/0.05\,\mathrm{s})$,
and an event-boundary flag; entries are zero where poses are invalid):
\begin{equation}
a_{t,h} \leftarrow a_{t,h} + \lambda_{\ell} \,\sigma\!\big(W_h g_t + b_h\big),
\qquad \sigma=\tanh,\quad \lambda_{\ell}=0,\ b_h=0\text{ at initialization},
\label{eq:gate}
\end{equation}
where $\lambda_{\ell}$ is one trainable scalar per layer and $W$ has a
random initialization (standard deviation 0.02). The tanh term is bounded;
the learned scalar is unconstrained. Thus the patch is an exact no-op at initialization
($6\cdot16+16+1=113$ parameters per layer, 2{,}034 total). The clock control
uses six normalized clock channels (three local and three global) with the
same affine--tanh--scalar form and parameter count; the shuffled-training
control permutes the valid timesteps of $g_t$ within each episode with a fixed
seed during training. Evaluation uses correct online geometry for both f0 and
f0-shuffled; LoRA excludes the decay projections, so decay-gate conditioning exists
only through Eq.~\eqref{eq:gate}.

\begin{table}[t]
\centering\small
\setlength{\tabcolsep}{5pt}
\caption{Condition matrix. Content: none / relative geometry (14-d: object- and goal-relative translations and an invariant axis block) / 6-d invariant increments $g_t$ / scrambled $g_t$ / token clock.}
\label{tab:cond}
\begin{tabular}{llllr}
\toprule
id & pathway & content & role & trainable\\
\midrule
c0 & --- & no object/goal geometry & baseline & 6{,}215{,}224\\
c1 & tokens & same $g_t$ via 6$\to$2 MLP & matched-input control & 6{,}217{,}286\\
c2 & tokens & relative geometry (14-d concat) & pathway A & 6{,}229{,}560\\
f0 & decay gate & invariant increments $g_t$ & pathway B & 6{,}217{,}258\\
f0-shuffled & decay gate & shuffled training $g_t$ & alignment control & 6{,}217{,}258\\
b1 & decay gate & token clock & content control & 6{,}217{,}258\\
\bottomrule
\end{tabular}
\end{table}

\paragraph{Matched-input pathway comparison.}
c1 embeds the same six-dimensional $g_t$ used by f0 through a 6$\to$2
linear layer and GELU, concatenated before the shared input projection.
Its total trainable count differs from f0 by 28 parameters. The raw inputs match, but the token bottleneck and layerwise gate adapter
transform them differently. c2 uses a separate 14-d state representation.

\paragraph{Protocol.} Three-task joint training (35 demos per task; split
manifest with SHA-256 provenance), three seeds, evaluation on ten held-out
initial states per task --- $6\times3\times3\times10=540$ rollouts --- with
stage decomposition (reach/grasp/transport/place) recorded per episode.
Training uses 3,000 optimizer steps and batch size 32. The baseline retains
proprioceptive inputs, previous actions, and language as well as vision;
``no geometry'' means no explicit object/goal geometry, not an image-only policy.
We report Wilson intervals and exact paired McNemar tests for descriptive
comparisons over identical seed/task/initial-state keys. These intervals and
$p$-values omit dependence from shared training seeds and repeated initial states. We additionally
report per-cell outcomes and exploratory per-seed effect sizes, leave-one-seed-out
ranges, and a sensitivity analysis excluding the near-floor task 8. The sensitivity analyses are exploratory; the historical ledger and analysis
provenance are described in Appendix~A. Training loss is only an optimization diagnostic.

%% file: sections/findings.tex
\subsection{Training-time alignment}
Table~\ref{tab:main} summarizes the 540 held-out rollouts. Correct gate
geometry (f0) scores 26/90, compared with 33/90 for shuffled training and
22/90 for c0. The f0--shuffled contrast is $-7.8$ percentage points (pp;
paired wins--losses 12--19, descriptive $p=0.28$). Correct temporal alignment provides no clear benefit in this comparison.
Both policies receive correct geometry at evaluation, so inference-time
reliance on geometry remains untested.

\paragraph{Matched-input pathways.}
The c1 token-MLP and f0 gate adapters receive the same $g_t$ and score
25/90 versus 26/90. Their difference is $-1.1$\,pp (13--14, descriptive
$p=1.0$), with per-seed differences of $-10.0$, $-6.7$, and $+13.3$\,pp.
The comparison is inconclusive, with a possible bottleneck from c1's
two-dimensional embedding.

The token configuration c2 improves over c0 by 10.0\,pp (16--7,
$p=0.093$). The sample is too small to resolve this potentially useful effect. The c2--f0 difference is 5.6\,pp (15--10, $p=0.42$); because
c2 carries 14-d relative state and f0 carries 6-d increments, this contrast
cannot identify a pathway effect. Shuffled training exceeds c0 by 12.2\,pp
(18--7, $p=0.043$, uncorrected); the nine cell-level signs are 6 wins,
1 tie, and 2 losses (descriptive sign-test $p=0.289$). The cell-level result weakens the evidence for an improvement.

\paragraph{Seed and task sensitivity.}
Table~\ref{tab:sensitivity} reports exploratory sensitivity analyses. The
f0--shuffled difference changes sign across seeds; omitting task 8 narrows
it from $-7.8$ to $-1.7$\,pp. The effect is sensitive to both seed and task selection.
The c2--c0 trend stays positive when any one seed is omitted, but one of
three individual seeds favors c0. Shuffled training could provide a
regularization effect, or both geometry variants could be limited by the
training recipe; the present results do not distinguish these explanations.

\begin{table}[t]
\centering\small
\caption{Held-out success (3 seeds $\times$ 3 tasks $\times$ 10 episodes)
and paired McNemar tests vs.\ c0 over identical initial states (exact seed/task/episode pairing; descriptive, see Sec.~2). $^\dagger$One b1 seed failed to converge (App.~B); excluding it, b1 vs.\ c0 is 3--8, $p{=}0.23$.}
\label{tab:main}
\begin{tabular}{lccc}
\toprule
condition & success & rate [Wilson interval] & vs.\ baseline (wins--losses, $p$)\\
\midrule
f0-shuffled (training) & 33/90 & 36.7 [27.4, 47.0] & 18--7, 0.043\\
c2 (tokens, relative) & 31/90 & 34.4 [25.4, 44.7] & 16--7, 0.093\\
f0 (gate, correct) & 26/90 & 28.9 [20.5, 39.0] & 14--10, 0.541\\
c1 (tokens, MLP) & 25/90 & 27.8 [19.6, 37.8] & 12--9, 0.664\\
c0 (no geometry) & 22/90 & 24.4 [16.7, 34.2] & ---\\
b1 (gate, clock) & 10/90 & 11.1 [6.1, 19.3] & 3--15, 0.0075$^\dagger$\\
\bottomrule
\end{tabular}
\end{table}

\begin{table}[t]
\centering\small
\caption{Exploratory success differences (percentage points). Each seed
contains 30 rollouts per condition. Leave-one-seed-out (LOSO) ranges are
sensitivity ranges, not confidence intervals. Task exclusion retains tasks 0/4.}
\label{tab:sensitivity}
\begin{tabular}{lrrrcr}
\toprule
contrast & seed 0 & seed 1 & seed 2 & LOSO range & without task 8\\
\midrule
c1 $-$ f0 & $-10.0$ & $-6.7$ & 13.3 & [$-8.3$, 3.3] & $-6.7$\\
c2 $-$ c0 & 20.0 & 16.7 & $-6.7$ & [5.0, 18.3] & 11.7\\
f0 $-$ shuffled & $-6.7$ & 6.7 & $-23.3$ & [$-15.0$, 0.0] & $-1.7$\\
f0 $-$ b1 & 36.7 & 16.7 & 0.0 & [8.3, 26.7] & 26.7\\
b1 $-$ c0 & $-23.3$ & $-6.7$ & $-10.0$ & [$-16.7$, $-8.3$] & $-20.0$\\
\bottomrule
\end{tabular}
\end{table}

\subsection{Clock conditioning}
The correct-geometry gate scores 28.9\%, compared with 11.1\% for the
clock gate (20--4, descriptive $p=0.0015$). Both use the decay-gate interface,
but their inputs differ in semantics, scale, and temporal statistics; this is
not a distribution-matched test of semantic content alone. One of three b1
seeds fails to converge (L1 loss 0.882, versus 0.262 and 0.258 for its
siblings). We retain that seed in the primary outcome because optimization
failure is relevant to adaptation reliability.

\begin{figure}[t]
\centering
\includegraphics[width=0.85\linewidth]{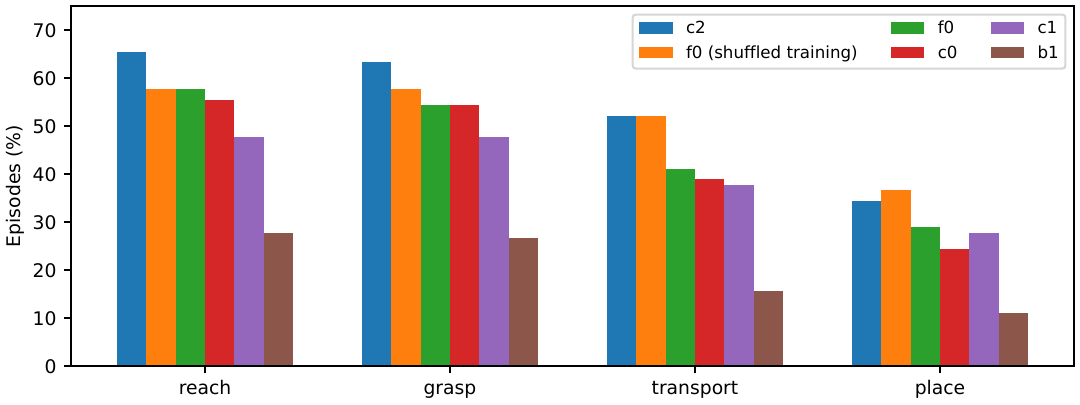}
\caption{Main-table stage rates across 90 episodes per condition. Reaching success is lower for b1 than for c0.}
\label{fig:stages}
\end{figure}

Excluding seed 0 from both conditions, b1 scores 10/60 (16.7\%) and c0
15/60 (25.0\%): b1--c0 is 3--8, $p=0.227$. The corresponding f0--b1
contrast is 9--4, $p=0.267$. The pooled deficit therefore combines an
optimization failure with smaller closed-loop differences among the remaining
seeds. Reaching declines from 55.6\% for c0 to 27.8\% for b1
(Fig.~\ref{fig:stages}).

A plausible hypothesis is that demonstration time predicts task phase,
encouraging a shortcut that fails when closed-loop timing drifts. Input scaling,
optimization sensitivity, and other explanations remain possible. Gate-activation
measurements and scale-matched controls would be needed to distinguish them.
More seeds are needed to estimate the frequency of training failures.

\subsection{Frame relabeling}
The frame experiment uses earlier state-only policies on task 0 and ten
training initial states, not the held-out visual checkpoints in Table~\ref{tab:main}.
A world-frame relabeling rotates the supplied poses by $\varphi$; world-frame
action translations are mapped back before simulator execution. The physical
scene is unchanged. Relative pose features are invariant under a common rigid
transform: $(GT_{\mathrm{ee}})^{-1}(GT_{\mathrm{obj}})
=T_{\mathrm{ee}}^{-1}T_{\mathrm{obj}}$. The relative features therefore remain unchanged without retraining.

\begin{figure}[t]
\centering
\includegraphics[width=\linewidth]{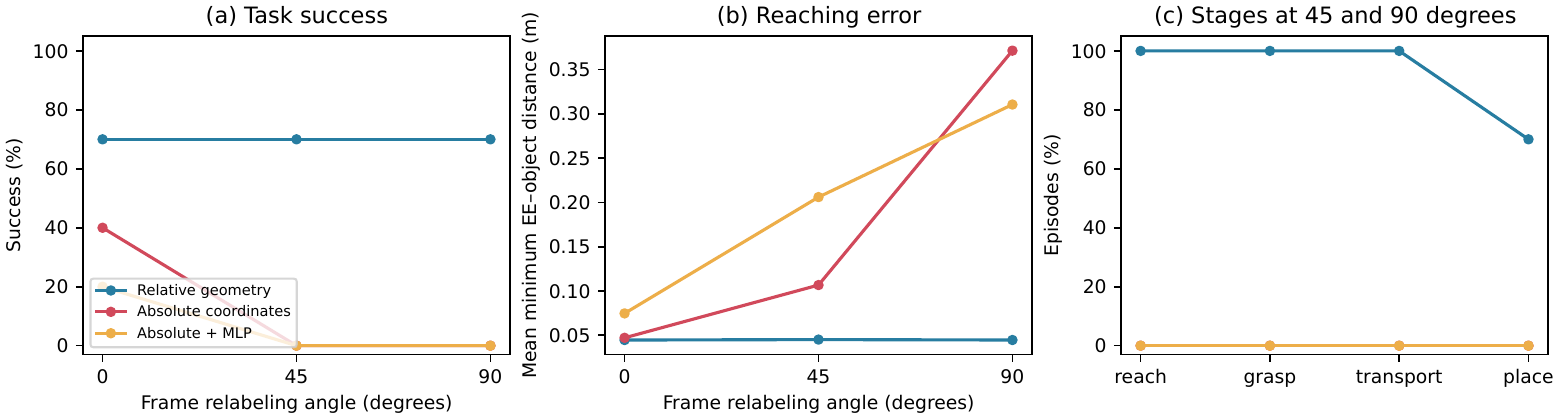}
\caption{State-only frame diagnostic on ten training initial states per angle.
(a) Success; (b) mean minimum reaching distance; (c) stage rates pooled over
$45^\circ$ and $90^\circ$. Values are recomputed from the supplied staged records.}
\label{fig:collapse}
\end{figure}

In the staged records supplied here, the absolute-coordinate policy scores
4/10, 0/10, and 0/10 at $0^\circ$, $45^\circ$, and $90^\circ$; its mean minimum
end-effector--object distance rises from 0.047 to 0.107 and 0.371\,m.
The relative policy scores 7/10 at all three angles, with 100\% reaching,
grasping, and transport (Fig.~\ref{fig:collapse}). An absolute-coordinate MLP
variant also fails at nonzero angles, but its 2/10 unrotated score makes it a
weak capacity control. The absolute policies are sensitive to frame changes, although the experiment
does not identify the mechanism of failure.

\subsection{Physical object displacement}
We separately displace the manipulated object by 5 and 10\,cm in the table
plane, using a deterministic per-episode diagonal direction. Table~\ref{tab:layout}
evaluates seed-0 visual checkpoints on tasks 0 and 4, with 20 held-out rollouts
per condition and displacement. Unlike frame relabeling, this changes the
physical task. The evaluations run on a second machine; its 0\,cm control
scores are reported alongside the shifted scores to provide a local reference.
They differ from the original main-table evaluations, so we do not pool machines.

All four tested policies have at most 3/20 success at 5\,cm and 0/20 at
10\,cm. Reaching falls from 70--85\% at 0\,cm to 15--55\% at 5\,cm and
5--10\% at 10\,cm. Geometry therefore does not prevent the observed collapse.
Distance errors are consistent with poor adaptation to displaced targets, but
minimum distances alone cannot distinguish layout memorization from other
control failures. Thus frame invariance alone is insufficient to predict performance after
physical displacement.

\begin{table}[t]
\centering\small
\caption{Physical layout shift (seed-0 checkpoints, tasks 0 and 4, 10
held-out episodes each): success / reach / mean minimum EE--object distance.}
\label{tab:layout}
\begin{tabular}{lccc|ccc|ccc}
\toprule
 & \multicolumn{3}{c|}{success /20} & \multicolumn{3}{c|}{reach \%} & \multicolumn{3}{c}{min EE--obj (m)}\\
condition & 0\,cm & 5\,cm & 10\,cm & 0\,cm & 5\,cm & 10\,cm & 0\,cm & 5\,cm & 10\,cm\\
\midrule
c0 (no geometry) & 7 & 1 & 0 & 70 & 30 & 10 & 0.048 & 0.061 & 0.096\\
c2 (tokens) & 12 & 1 & 0 & 80 & 50 & 5 & 0.054 & 0.069 & 0.102\\
f0 (gate) & 12 & 1 & 0 & 70 & 15 & 10 & 0.056 & 0.075 & 0.099\\
f0-shuffled & 15 & 3 & 0 & 85 & 55 & 5 & 0.050 & 0.062 & 0.094\\
\bottomrule
\end{tabular}
\end{table}

%% file: sections/discussion.tex
The shuffled-training control changes how the geometry result should be
read: f0 improves over c0 in aggregate, but correct temporal alignment does
not improve over shuffled training. A test-time gate knockout would answer
the separate question of whether either trained policy relies on geometry.
The clock result also depends on training reliability: its pooled deficit
shrinks when the failed seed is excluded. Finally, the frame and displacement
tests measure different properties and should be reported separately.

\paragraph{Limitations.}
We evaluate one model, three tasks, and three training seeds in simulation.
The 24.4\% baseline and near-floor task 8 leave open the possibility that a
stronger training recipe would reveal geometry benefits. The matched-input
c1 adapter has a two-dimensional bottleneck; wider embeddings could change
the pathway comparison. The frame test uses state-only checkpoints and
training initial states, while the layout test covers one seed and two tasks.
Closed-loop outcomes also vary across machines. We measure adaptation size,
but not on-device latency, memory consumption, or energy.

%% file: sections/appendix.tex
\section{Reproducibility and analysis provenance}
The ancillary file \texttt{supplementary.zip} contains 540 main-table, 90 staged frame-relabeling, and 240
layout-shift episode records, the split manifest, and the historical experiment
ledger. The script \texttt{scripts/analyze.py} recomputes success counts,
descriptive Wilson intervals, paired tests, cell signs, per-seed and task-exclusion
sensitivity analyses, stage rates, and frame/layout distance summaries.
The optional \texttt{--figures DIR} argument regenerates the two rollout figures with
Matplotlib. The architecture diagram is provided as editable TikZ source. Duplicate episode keys and incomplete main-table pairings cause
an error. Training summaries provide the reported fit losses and parameter
counts; these are distinct from rollout outcomes.

The ledger is a chronological research record containing planned gates,
protocol changes, and observations, not an immutable registration of the
final manuscript. The present sensitivity analyses and revised interpretations
were added after inspecting the outcomes. Earlier pooled statistics in the
historical ledger are superseded by the executable analysis with exact
seed/task/episode pairing. The current figures use the supplied staged records
consistently; no cross-run selection is performed.

\section{Training-fit table (optimization diagnostic only)}
Means $\pm$ sd over 3 seeds (L1): f0-shuffled $0.214\pm0.019$; f0
$0.217\pm0.001$ (training-loss variability only); c1 $0.256\pm0.010$; c0 $0.282\pm0.087$; c2 $0.295\pm0.065$; b1 seeds $\{0.882, 0.262, 0.258\}$ --- seed~0 failed to converge, and its closed-loop score (0/30) is reported both pooled and excluded in Sec.~3.2. Fit ordering is recipe-dependent and does not predict
closed-loop success --- another argument for closed-loop-first evaluation.

\section{Per-cell success table}
\label{app:cells}
Successes out of 10 held-out episodes per (seed, task) cell; tasks 0/4/8 of
LIBERO-Spatial.
\begin{center}\small
\begin{tabular}{lccc|ccc|ccc|c}
\toprule
 & \multicolumn{3}{c|}{seed 0} & \multicolumn{3}{c|}{seed 1} & \multicolumn{3}{c|}{seed 2} & \\
condition & t0 & t4 & t8 & t0 & t4 & t8 & t0 & t4 & t8 & total\\
\midrule
c0 & 3 & 4 & 0 & 7 & 0 & 0 & 8 & 0 & 0 & 22\\
c1 & 2 & 4 & 2 & 8 & 0 & 0 & 7 & 1 & 1 & 25\\
c2 & 6 & 5 & 2 & 4 & 8 & 0 & 6 & 0 & 0 & 31\\
f0 & 6 & 5 & 0 & 5 & 5 & 0 & 4 & 1 & 0 & 26\\
f0-shuffled & 7 & 2 & 4 & 6 & 2 & 0 & 9 & 1 & 2 & 33\\
b1 & 0 & 0 & 0 & 5 & 0 & 0 & 3 & 2 & 0 & 10\\
\bottomrule
\end{tabular}
\end{center}
Task 8 is near the floor for every condition; seed-to-seed variation within a
condition (e.g.\ c2: 13/12/6 per seed) is of the same order as the
between-condition differences in Table~\ref{tab:main}, which is why we treat the paired
tests as descriptive.

\section{Research practice and broader impacts}
\label{app:practice}
Qwen3.5-0.8B is the policy backbone described in Sec.~2. AI coding assistants
also supported experimental planning, implementation and debugging, analysis
scripts, and manuscript preparation. The reported success rates are computed
from recorded simulator outcomes; language-model judgments are not used to
score rollouts. The revised analyses are identified as exploratory in Appendix~A.

The failure cases may help researchers avoid unsupported claims of spatial
robustness in small robot policies. Deploying such policies on physical robots
without further validation could cause collisions or property damage,
particularly after scene changes. The experiments reported here are confined
to simulation and do not establish safe physical deployment. No new human-subject
study or crowdsourced annotation was conducted for these experiments.

The supplement supports reanalysis of the recorded rollouts. It does not
include a complete training/evaluation environment or the trained checkpoints,
and a complete accounting of per-run and exploratory compute is unavailable.
The model and benchmark are credited in the paper; a full dependency and
asset-license inventory is not included in this submission.